\documentclass[10pt,twocolumn,letterpaper]{article}
\pdfoutput=1

\usepackage{iccv}
\usepackage{times}
\usepackage{epsfig}
\usepackage{amsmath}
\usepackage{amssymb}

\usepackage{amsmath,amsfonts,bm}

\def\eqref#1{equation~\ref{#1}}

\def\1{\bm{1}}

\DeclareMathAlphabet{\mathsfit}{\encodingdefault}{\sfdefault}{m}{sl}
\SetMathAlphabet{\mathsfit}{bold}{\encodingdefault}{\sfdefault}{bx}{n}

\usepackage[utf8]{inputenc} 
\usepackage[T1]{fontenc}    
\usepackage{url}            
\usepackage{booktabs}       
\usepackage{amsfonts}       
\usepackage{nicefrac}       
\usepackage{microtype}      
\usepackage{multirow}
\usepackage{subfigure}
\usepackage{booktabs}
\usepackage{xcolor}
\usepackage{caption}
\usepackage{epsfig}
\usepackage{verbatim}
\usepackage{authorname}
\usepackage{graphicx}

\usepackage[pagebackref=true,breaklinks=true,letterpaper=true,colorlinks,bookmarks=false]{hyperref}

\iccvfinalcopy 

\def\iccvPaperID{****} 

\ificcvfinal\fi

\begin{document}

\title{MeshMVS: Multi-View Stereo Guided Mesh Reconstruction}

\author{
    \small
    Rakesh Shrestha$^{1}$, Zhiwen Fan$^{2}$, Qingkun Su$^{2}$, Zuozhuo Dai$^{2}$, Siyu Zhu$^{2}$, Ping Tan$^{1}$ \\
    \normalsize{Simon Fraser University$^{1}$},
    \normalsize{Alibaba A.I Labs$^{2}$} \\
    \scriptsize{\{rakeshs,pingtan\}@sfu.ca, \{waynefan.fzw,siting.zsy,qingkun.sqk,zuozhuo.dzz\}@alibaba-inc.com}
}

\maketitle
\ificcvfinal\thispagestyle{empty}\fi

\newcommand{\todo}[1]{{\textcolor{red}{\bf [#1]}}}

\newcommand{\figref}[1]{Figure~\ref{fig:#1}}
\newcommand{\tabref}[1]{Table~\ref{tab:#1}}
\newcommand{\equref}[1]{Equation~(\ref{equ:#1})}
\newcommand{\secref}[1]{Section~\ref{sec:#1}}
\newcommand{\subsecref}[1]{Sub-section~\ref{subsec:#1}}
\newcommand{\tableref}[1]{Table~\ref{table:#1}}

\begin{abstract}
    Deep learning based 3D shape generation methods generally utilize latent features extracted from color images to encode the semantics of objects and guide the shape generation process.
    These color image semantics only implicitly encode 3D information, potentially limiting the accuracy of the generated shapes.
    In this paper we propose a multi-view mesh generation method which incorporates geometry information explicitly by using the features from intermediate depth representations of multi-view stereo and regularizing the 3D shapes against these depth images.
    First, our system predicts a coarse 3D volume from the color images by probabilistically merging voxel occupancy grids from the prediction of individual views.
    Then the depth images from multi-view stereo along with the rendered depth images of the coarse shape are used as a contrastive input whose features guide the refinement of the coarse shape through a series of graph convolution networks.
    Notably, we achieve superior results than state-of-the-art multi-view shape generation methods with 34\% decrease in Chamfer distance to ground truth and 14\% increase in F1-score on ShapeNet dataset.
    Our source code is available at \url{https://git.io/Jmalg}


\end{abstract}

\section{Introduction}

3D shape generation is a long-standing research problem in computer vision and computer graphics with applications in autonomous driving, augmented reality, etc. Conventional approaches mainly leverage multi-view geometry based on stereo correspondences between images but are restricted by the coverage provided by the input views. With the availability of large-scale 3D shape datasets and the success of deep learning in several computer vision tasks, 3D representations such as voxel grid~\cite{3dr2n2, tulsiani2017multi, yan2016perspective} and point cloud~\cite{yang2018foldingnet, fan2017point} have been explored for single-view 3D reconstruction.
Among them, triangle mesh representation has received the most attention as it has various desirable properties for a wide range of applications and is capable of modeling detailed geometry without high memory requirement.

Single-view 3D reconstruction methods~\cite{wang2018pixel2mesh,huang2015single,kar2015category,su2014estimating} generate the 3D shape from merely a single color image but suffer from limited visibility due to occlusion and high dependency on training views leading to weaker generalization to different input views, resulting in low quality reconstructions.
Multi-view methods~\cite{wen2019pixel2mesh++,3dr2n2,kar2017lsm,mcrecon2017} extend the input to images from different viewpoints which provides more visual information and improves the accuracy of the generated shapes.
Recent work in multi-view mesh reconstruction~\cite{wen2019pixel2mesh++} introduces a multi-view deformation network using perceptual feature from each color image for refining the meshes generated by Pixel2Mesh~\cite{wang2018pixel2mesh}.
Although promising results were obtained, this method relies on perceptual features from color images which do not explicitly encode the geometry of the objects and could restrict the accuracy of the 3D models.
The work is also constrained by the topology of the initial shape, an ellipsoid, which limits the accuracy of the shapes from subsequent refinement modules.



In this work, we present a novel multi-view mesh generation method where we start by predicting coarse volumetric occupancy grid representations of the color images.
Using this representation for further refinement rather than a predefined template shape allows us to generate shapes with more accurate topology~\cite{gkioxari2019meshrcnn}.
We then use Graph Convolutional Network (GCN)~\cite{scarselli2008graph,wang2018pixel2mesh} to fine-tune the volumetric representation to surface mesh in a coarse-to-fine manner.
The GCN obtains features of the graph nodes (mesh vertices) from contrastive depth features alongside the RGB perceptual features.
The contrastive depth features are extracted from the rendered depth maps of the intermediate shape and predicted depth maps from a multi-view stereo network.
Constrains between the rendered depths and predicted depths at different viewpoints are further added.
This allows our system to better reason about the transformation required to deform the intermediate shapes to confirm to the predictions by the multi-view stereo network.

Both qualitative and quantitative experimental results based on the ShapeNet~\cite{chang2015shapenet} benchmark are provided to demonstrate the effectiveness of the proposed approach. Remarkably, our method achieves the best performance among all the previous mesh generation methods, with 14\% increase in F1-score compared to the state-of-the-art.



\section{Related Work}
In this section, we first focus on the representation issue of 3D deep learning. Afterwards, a brief review of shape generation approaches based on single view and multiple views is represented respectively.

\paragraph{3D Shape Representation}\vspace{-4mm}
3D occupancy grid has been a popular representation since conventional CNNs can be used to generate them~\cite{3dr2n2,kar2017lsm}.
More recently, mesh representation is being increasingly used for 3D reconstruction~\cite{wang2018pixel2mesh,wen2019pixel2mesh++} where a template mesh (an ellipsoid) is deformed to obtain the final shape.
This approach struggles in reconstructing shapes whose topology is very different from the template mesh.
\cite{gkioxari2019meshrcnn} proposes a hybrid approach where a coarse occupancy grid is first predicted from single image which is converted to mesh representation and further refined using mesh deformations.
We adopt a similar approach in our work where we predict the coarse occupancy grid from multi-view images which are used as the initial shape for succeeding refinements.
Other shape representations include point clouds~\cite{fan2017point,yang2018foldingnet,jia2020dv}, implicit surfaces~\cite{park2019deepsdf}, depth images~\cite{yao2018mvsnet,yao2019recurrent} etc.

\begin{figure*}[t]
\begin{center}
\includegraphics[width=0.95\linewidth]{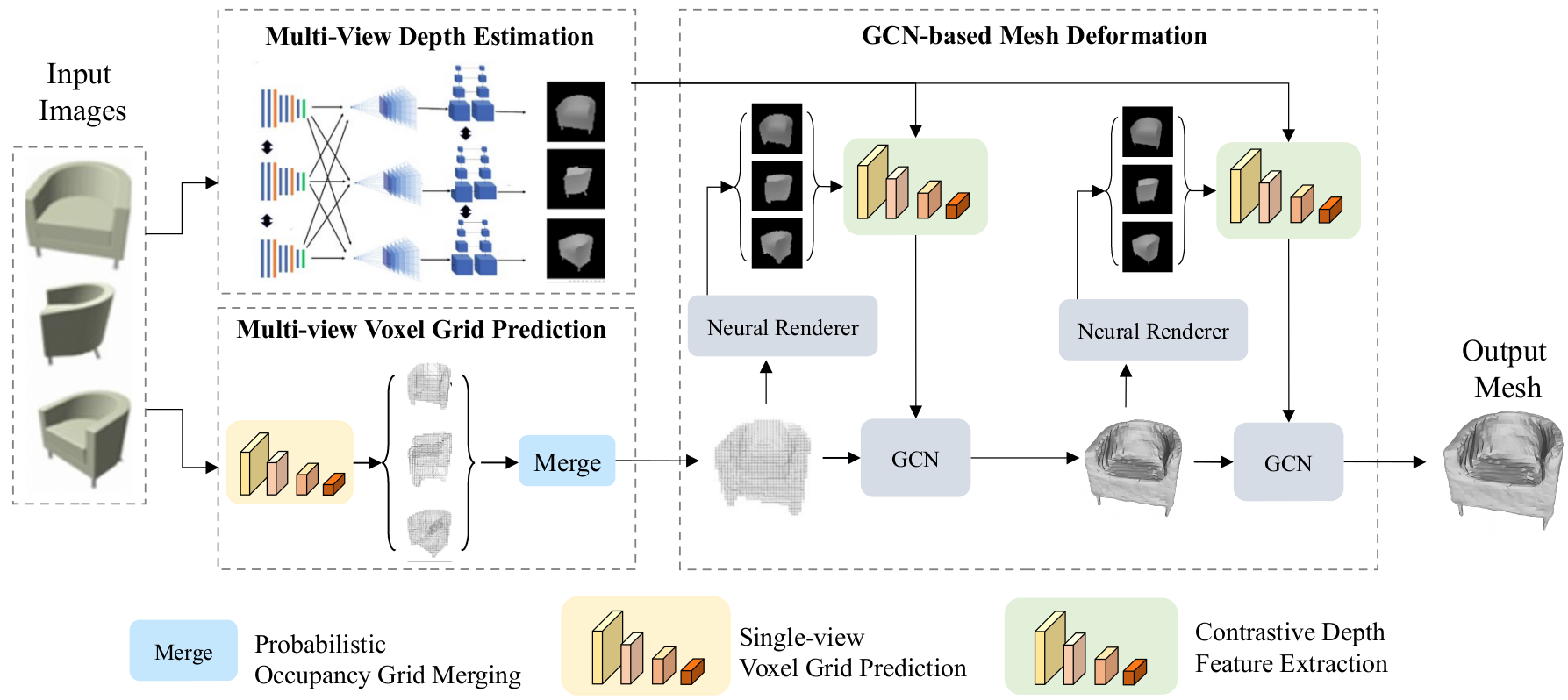}
\end{center}
\caption{
    \textbf{Architecture of the proposed method}.
    First, the voxel grid prediction module obtains a coarse voxel grid representation by probabilistically merging predicted voxel grids of each input image.
    Then, a series of GCNs further refine the cubified voxel grid in a coarse-to-fine manner using contrastive features from the rendered depths of the current shape and the depths from multi-view stereo.
    Specifically, the multi-view features are pooled using a attention-based mechanism.
}
\label{fig:system_architecture}
\end{figure*}

\paragraph{Single-view Shape Generation.}\vspace{-4mm}
Traditional single-view shape generation methods like~\cite{durou2008numerical,zhang1999shape,favaro2005geometric} reason about shading, texture and defocus to reason about visible parts of the object and infer its 3D geometry.
Earlier learning-based approaches~\cite{huang2015single, su2014estimating} use shape component retrieval and deformation from a large dataset for single-view 3D shape generation.
\cite{kurenkov2018deformnet} extend this idea by introducing free-form deformation networks on retrieved object templates from a database.
Some work learn shape deformation from ground truth foreground masks of 2D images~\cite{kar2015category,yan2016perspective,tulsiani2017multi}.
\cite{3dr2n2,hane2017hierarchical,johnston2017scaling} can learning 3D volumetric representations through deep learning.
Predicting 3D mesh from single-view color images has been proposed in~\cite{wang2018pixel2mesh,pan2019deep,gkioxari2019meshrcnn, tang2019skeleton}.
DR-KFS~\cite{jin2019drkfs} introduces a differentiable visual similarity metric to learn single-view 3D shape generation
while SeqXY2SeqZ~\cite{han2020seqxy2seqz} represents 3D shapes using a set of 2D voxel tubes for shape reconstruction.
Front2Back~\cite{yao2020front2back} generates 3D shapes by fusing predicted depth and normal images and
DV-Net~\cite{jia2020dv} predicts dense object point clouds using dual-view RGB images with a gated control network to fuse point clouds from the two views.
FoldingNet~\cite{yang2018foldingnet} learns to reconstruct arbitrary point clouds from a single 2D grid.
AtlasNet~\cite{groueix2018papier} use learned parametric representation
while \cite{mescheder2019occupancy,park2019deepsdf,liu2019learning,liu2019dist,murez2020atlas} employ implicit surface representation to reconstruct 3D shapes.

\paragraph{Multi-view Shape Generation.}\vspace{-4mm}
Multi-view 3D model generation has traditionally been tackled using stereo geometry principles.
Among them, structure-from-motion (SfM)~\cite{schonberger2016structure,agarwal2011building,cui2015global,cui2017hsfm} and simultaneous localization and mapping (SLAM)~\cite{cadena2016pastslam,mur2015orb,engel2014lsd,whelan2015elasticfusion} are popular techniques that perform 3D reconstruction and camera pose estimation at the same time.
Similarly, traditional multi-view stereo methods infer 3D geometry from images with known camera parameters either using
volumetric representation~\cite{kutulakos2000theory, seitz1999photorealistic} or
point cloud representation~\cite{furukawa2009accurate, lhuillier2005quasi}.
These methods extract local image features, match them across images and use the matches to estimate 3D geometry.
While the results of these works are impressive in terms of quality and completeness of reconstruction, they still struggle with poorly textured and reflective surfaces and require carefully selected input views.

Deep learning based approaches can learn to infer 3D structure from training data and can be robust against poorly textured and reflective surfaces as well as limited and arbitrarily selected input views.
\cite{hartmann2017learned_16,deepmvs2018,yao2018mvsnet,chen2019point,luo2019pmvsnet,gu2019cascade,yao2019recurrent} propose multi-view stereo models from learned cost volumes to predict depth images.
Recurrent Neural Networks (RNN) based methods~\cite{3dr2n2, kar2017lsm, mcrecon2017} are another popular solution to solve this problem.
\cite{mcrecon2017, lin2019photometric} introduce image silhouettes along with adversarial multi-view constraints and optimize object mesh models using multi-view photometric constraints.
Pixel2Mesh++~\cite{wen2019pixel2mesh++} extends single-view Pixel2Mesh~\cite{wang2018pixel2mesh} to multi-view image input by introducing cross-view perceptual feature pooling and multi-view deformation reasoning but struggles to reconstruct complex shape topology due to its use of ellipsoidal template shape which it deforms to the final shape.
Our work avoids this pitfall by first predicting a coarse volumetric model which can represent complex topology and then applying deformations on it to get a finer shape as the final model.

\section{Methodology}
\figref{system_architecture} shows the architecture of the proposed system which takes as input multi-view color images of an object with known poses and outputs a triangle mesh of the object $ T = (V, F) $ where $ V = \{ v_i \in \mathbb{R}^3\} $ is the set of vertex coordinates and $ F \subseteq V \times V \times V $ is the set of faces.
The multi-view voxel grid prediction module (\subsecref{multiview_voxel}) first predicts a coarse voxel grid of the object.
The voxel grid is converted to mesh representation by cubify operation~\cite{gkioxari2019meshrcnn} which is further refined by the mesh refinement module (\subsecref{mesh_refinement}).
A series of Graph Convolution Networks (GCN) deforms the cubified voxel grid to obtain surface mesh.
The GCNs use multi-view contrastive depth features from concatenated predicted and rendered depth along with RGB features as input to deform the input mesh.
Depth images are predicted using an extended MVSNet network~\cite{yao2018mvsnet}.
The multi-view features are fused using attention mechanism.

\subsection{Multi-view Voxel Grid Prediction}
\label{subsec:multiview_voxel}

\paragraph{Single-view Voxel Grid Prediction}
For each of the input single RGB image, we first generate the voxel occupancy grid of the target object.
Specifically, the single-view voxel branch adopts the approach proposed in~\cite{gkioxari2019meshrcnn} which consists of a ResNet feature extractor and a fully convolutional voxel grid prediction network.
Here, we set the resolution of the generated voxel occupancy grid as 48 $\times$ 48 $\times$ 48.
The voxel prediction networks for all viewpoints share the same weights.

\paragraph{Probabilistic Occupancy Grid Merging}\vspace{-4mm}
Voxel occupancy grid predicted from a single viewpoint suffers from occlusion and limited visibility.
In order to fuse voxel grids from different viewpoints we take inspiration from occupancy map building techniques in robotics~\cite{grisetti2007improved,konolige1997improved} and propose a probabilistic occupancy grid merging method which merges the voxel grids from each input viewpoint probabilistically to obtain the final voxel grid output.
This allows occluded regions in one view to be estimated from other views where those regions are visible as well as increase the confidence of prediction in overlapping regions.
Occupancy probability of each voxel is represented by $p(x)$ which is converted to log-odds (logit):

\begin{equation}
    l(x) = log \frac{p(x)}{1 - p(x)}
    \label{equ:logodds}
\end{equation}

\begin{figure}[th!]
    \begin{center}
        \includegraphics[width=\linewidth]{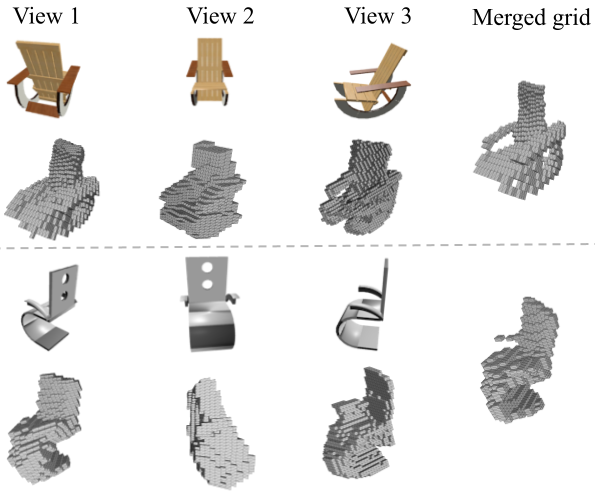}
    \end{center}
        \caption{\textbf{Two examples of voxel grid prediction of individual views along with probabilistically merged grid}. For each example, top row: input images, bottom row: single-view voxel grid prediction of the corresponding images transformed to the same coordinate frame. The merged grid can take into account the visibility limitations in single view predictions and produce better quality initial shape.}
        \label{fig:initial_vox}
\end{figure}

Bayesian update on the probabilities reduce to simple summation of log likelihoods~\cite{konolige1997improved}. Hence, the multi-view log-odds of a voxel is given by:

\begin{equation}
    l(x) = l_1(x) + l_2(x) + ... + l_n(x)
    \label{equ:logodds_sum}
\end{equation}

\noindent where $l_i$ is the voxel's log-odds in view $i$ and $n$ is the number of input views.
The final voxel probability $x$ is obtained by applying the inverse function of \equref{logodds} which is a sigmoid function.
\figref{initial_vox} shows the voxel grid predictions at each viewpoint along with the merged grid for two sets of input images.

\subsection{Mesh Refinement}
\label{subsec:mesh_refinement}
The cubified mesh from the voxel branch only provides a coarse reconstruction of the object's surface.
We then apply the graph convolutional network in which the mesh vertex and edges are represented by the graph nodes and edges respectively, to deform the mesh to more accurate positions based on the RGB and contrastive depth features.
\label{subsec:depth_prediction}
\paragraph{Multi-View Depth Estimation}\vspace{-4mm}
We extend MVSNet~\cite{yao2018mvsnet} for multi-view depth prediction.
To obtain the depth image of each RGB image, we predict feature volume for each image using shared CNN and iteratively warp the features to each view and apply cost volume regularization to predict the depths.
In implementation, we discard the policy of view selection, and set the number of depth hypothesis to $48$ which is equivalent to a resolution of $25$ mm.




\paragraph{GCN-based Mesh Deformation}\vspace{-4mm}
Following the prestigious previous works~\cite{wang2018pixel2mesh,wen2019pixel2mesh++}, a graph convolution deforms mesh vertices by propagating features from neighboring vertices by applying:
\begin{equation}
f_{i}^{'} = ReLU(W_0f_i + \sum_{j \in \mathcal{N}(i)} W_1 f_j),
\end{equation}
where $\mathcal{N}(i)$ is the set of neighboring vertices of the \emph{i}-th vertex in the mesh, $f_{\{\}}$ represents the feature vector of a vertex, and $W_0$ and $W_1$ are learnable parameters of the model.
The features pooled from multi-view images (discussed in subsequent paragraphs) along with 3D coordinates of the vertices in world frame are used as features of the graph nodes.
Series of Graph-based Convolutional Network (GCN) blocks are applied to deform a mesh at the current stage to the next stage, starting with the cubified voxel grids.
Each GCN block utilizes several graph convolutions to transform the vertex features along with a final vertex refinement operation where the features along with vertex coordinates are further transformed as:
\begin{equation}
v_i^{'} = v_i + tanh(W_{vert}[f_i;v_i]).
\end{equation}
Here, $v_i$ and $v_i^{'}$ represent the vertex coordinates before and after each refinement operation, and the matrix $W_{vert}$ is another learnable parameter to obtain the deformed mesh.

\label{subsec:contrastive_depth_feature_extraction}
\paragraph{Contrastive Depth Feature Extraction}\vspace{-4mm}
\begin{figure}[th!]
    \begin{center}
        \includegraphics[width=\linewidth]{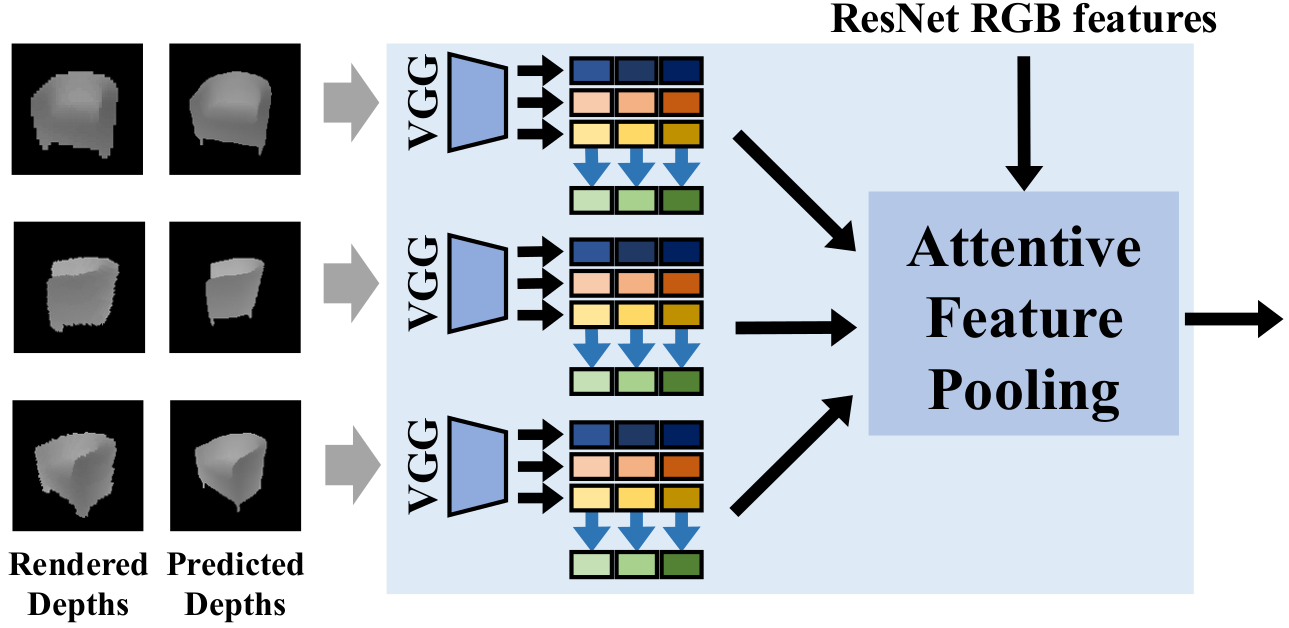}
    \end{center}
        \caption{\textbf{Contrastive feature extraction}. We concatenate rendered depths and predicted depths and apply CNN feature extractor. The multi-view features are pooled using attentive feature pooling and used by GCN for deforming the shape}
        \label{fig:contrastive_feature_extractor}
\end{figure}

We take inspiration from the work~\cite{yao2020front2back} that incorporates intermediate and image-centric 2.5D representations instead of directly generating 3D shapes from only RGB images~\cite{wang2018pixel2mesh,wen2019pixel2mesh++} to improve 3D reconstruction quality.
As shown in Figure~\ref{fig:contrastive_feature_extractor}, we therefore propose to formulate the features of graph nodes using depth images as the additional inputs alongside with the RGB features.
At a single viewpoint, we consider two sources of depth images, namely the rendered depth from the meshes at different GCN stages and the corresponding input depth from multi-view stereo.
We apply the approach proposed in~\cite{kato2018renderer} for differentiable rendering.
Afterwards, we propose a contrastive feature extractor which contrasts the rendered depth of the current mesh against the input depth from multi-view stereo and guides the GCN to reason about the deformation.
We use VGG-16~\cite{simonyan2014vgg} as our contrastive depth feature extraction network.
Given the 2D feature maps, the feature vector of each graph node can be obtained by projecting its 3D coordinate to the corresponding feature map using known camera parameters.

\paragraph{Attention-based Multi-View Feature Pooling}\vspace{-4mm}
In order to fuse multi-view contrastive depth features, we formulate an attention module by adapting multi-head attention mechanism originally designed for sequence to sequence machine translation using transformer architecture~\cite{vaswani2017attention}.
We choose multi-head attention as our feature pooling method since it allows the model to attend information from different representation subspaces of the features by training multiple attentions in parallel.
This method is also invariant to the order and number of input views.

\subsection{Loss functions}
\label{subsec:losses}

\paragraph{Mesh losses}
To constrain the mesh predicted by each GCN block $P$ to resemble the ground truth $Q$, we use three mesh loss functions the same as~\cite{wang2018pixel2mesh}.
The first loss measures the Chamfer distance~\cite{fan2017point} between the nearest neighbor mesh vertices $\Lambda$, and deforms vertices to correct positions.
It is defined as:

\begin{footnotesize}
\begin{equation}
\mathcal{L}_{\text{chamfer}}(P, Q) = |P|^{-1} \!\!\!\!\!\!\!\!\sum_{(p, q) \in \Lambda_{P,Q}}\!\!\!\!\!\!\!\!{||p-q||^{2}} + |Q|^{-1} \!\!\!\!\!\!\!\!\sum_{(q, p) \in \Lambda_{Q,P}}\!\!\!\!\!\!\!\!{||q-p||^{2}},
\end{equation}
\end{footnotesize}

\noindent Furthermore, to encode the local higher order surface properties, the normal loss function is defined as:

\begin{footnotesize}
\begin{equation}
\mathcal{L}_{\text{normal}}(P, Q) = -|P|^{-1}\!\!\!\!\!\!\!\! \sum_{(p, q) \in \Lambda_{P,Q}}\!\!\!\!\!\!\!\!{|u_p \cdot u_q|} - |Q|^{-1}\!\!\!\!\!\!\!\! \sum_{(q, p) \in \Lambda_{Q,P}}\!\!\!\!\!\!\!\!{|u_q \cdot u_p|},
\end{equation}
\end{footnotesize}

\noindent where $u_p$ and $u_q$ represent the normals of vertices $p$ and $q$ respectively.

An additional regularization term in the form of edge length loss is also applied which penalizes long edges resulting in flying vertices~\cite{wang2018pixel2mesh} for visually appealing results. This loss function is defined as:

\begin{equation}
\mathcal{L}_{\text{edge}}(V, E) = \frac{1}{|E|} \sum_{(v,v') \in E}{||v - v'||^2},
\end{equation}

\noindent where $E$ denotes the set of graph edges.


\paragraph{Multi-view stereo depth loss}\vspace{-4mm}
The original MVSNet~\cite{yao2018mvsnet} uses L1-loss, but this approach uses BerHu loss since it gives slightly higher accuracy.
An intuitive explanation is that BerHu provides a good balance between L1 and L2 loss and a similar improvement is shown in the work~\cite{laina2016deeper} as well.
The multi-view stereo depth loss is defined as:

\begin{equation}
  \mathcal{L}_{depth}=\begin{cases}
    |x|, & \text{if $|x| \le c$},\\
    \frac{x^2 + c^2}{2c}, & \text{otherwise}.
  \end{cases}
\end{equation}

\noindent Here, $x$ is the depth error of a pixel between the predicted and the ground-truth and $c=0.2$ is a constant.

\paragraph{Contrastive depth loss}\vspace{-4mm}
The BerHu loss is also applied to measure the difference between the rendered depth images at different GCN stages and the depth images from multi-view stereo.
We denote the constrastive depth loss as $\mathcal{L}_{contrastive}$.
Such a loss function encourages the predicted mesh to conform to the multi-view stereo depths.

\paragraph{Voxel occupancy loss}\vspace{-4mm}
We introduce the voxel occupancy loss which measures the cross-entropy loss between the predicted voxel occupancy probabilities and the ground truth occupancy.
Such a loss is used to supervise the voxel predictions and further constrain the coarse initial shapes~\cite{gkioxari2019meshrcnn}.
It is defined as:
\begin{small}
\begin{equation}
\mathcal{L}_{\text{voxel}} = -{\Big(p(x) log\big(p(x)\big) + \big(1 - p(x)\big)log\big(1 - p(x)\big)\Big)}
\end{equation}
\end{small}

\vspace{-4mm}
Finally, we use the weighted sum of the individual losses discussed above as the final loss and train our model in an end-to-end fashion.
The final loss term $\mathcal{L}$ is defined as:
$\mathcal{L} = \lambda_{\text{chamfer}}\mathcal{L}_{\text{chamfer}} + \lambda_{\text{normal}}\mathcal{L}_{\text{normal}} + \lambda_{\text{edge}}\mathcal{L}_{\text{edge}} + \lambda_{\text{depth}}\mathcal{L}_{\text{depth}} + \lambda_{\text{contrastive}}\mathcal{L}_{\text{contrastive}} + \lambda_{\text{voxel}}\mathcal{L}_{\text{voxel}}$.

\section{Experiments}
\begin{figure*}[t]
\begin{center}
\includegraphics[width=\linewidth]{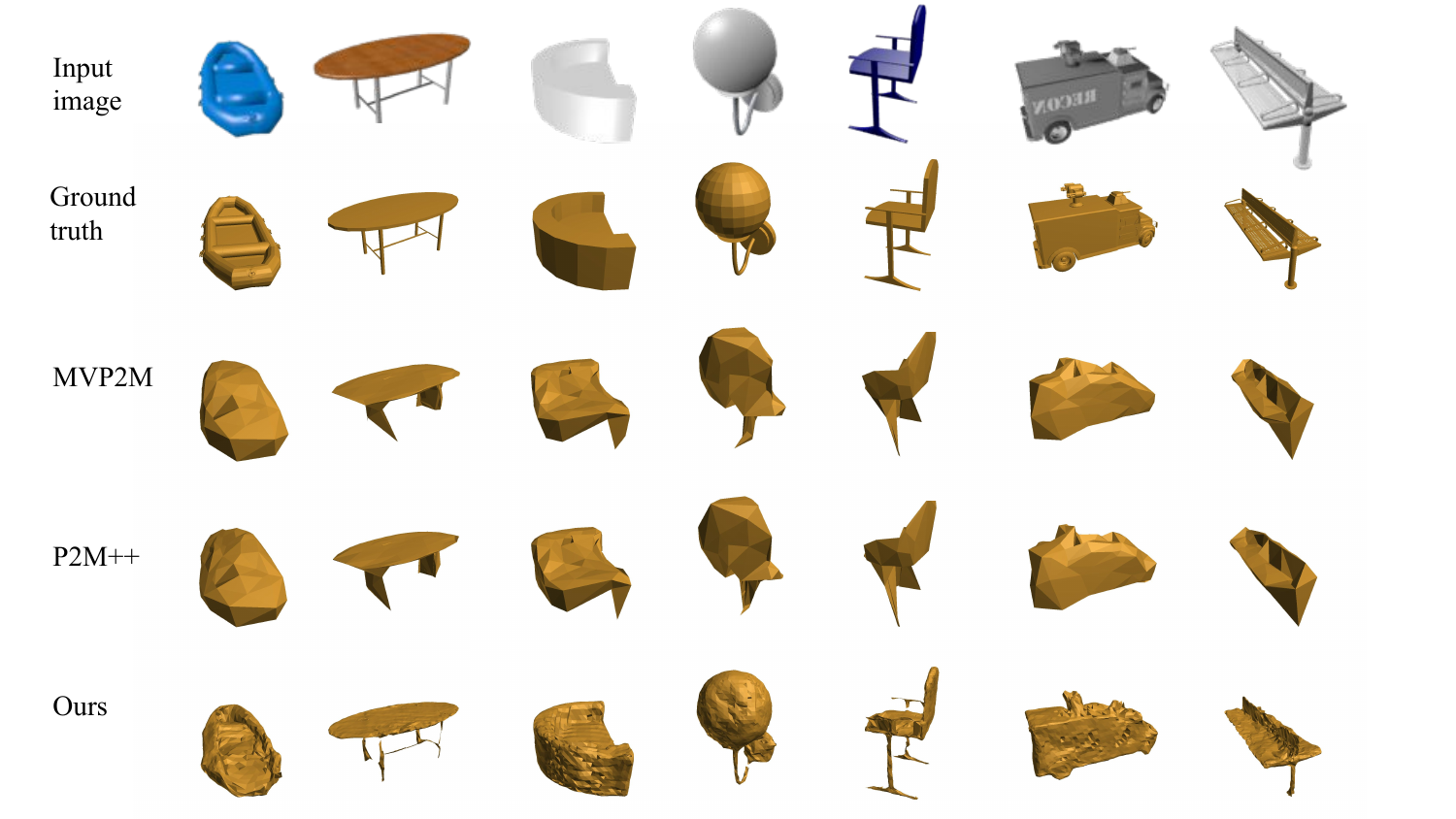}
\end{center}
\caption{
    \textbf{Qualitative evaluation} on ShapeNet dataset. \textbf{From top to bottom}: one of the input images, ground truth mesh, multi-view extended Pixel2Mesh, Pixel2Mesh++, and ours.
    Our predictions are closer to the actual shape, especially for the objects with more complex topology.}

\label{fig:qualitative_evaluation}
\end{figure*}

\subsection{Experimental Setup}
\label{subsec:experimental_setup}

\paragraph{Comparisons}
We evaluate the proposed method against various multi-view shape generation methods.
The state-of-the-art method is Pixel2Mesh++~\cite{wen2019pixel2mesh++} (referred as \emph{P2M++}). \cite{wen2019pixel2mesh++} also provide a baseline by directly extending Pixel2Mesh~\cite{wang2018pixel2mesh} to operate on multi-view images (referred as \emph{MVP2M}) using their statistical feature pooling method to aggregate features from multiple color images.
Results from additional multi-view shape generation baselines 3D-R2N2~\cite{3dr2n2} and LSM~\cite{kar2017lsm} are also reported.

\paragraph{Dataset}\vspace{-4mm}
We evaluate our method against the state-of-the-art methods on the dataset from~\cite{3dr2n2}
which is a subset of ShapeNet~\cite{chang2015shapenet} and has been widely used by recent 3D shape generation methods.
It contains 50K 3D CAD models from 13 categories.
Each model is rendered with a transparent background from 24 randomly chosen camera viewpoints to obtain color images.
The corresponding camera intrinsics and extrinsics are provided in the dataset.
Since the dataset does not contain depth images, we render them using a custom depth renderer at the same viewpoints as the color images and with the same camera intrinsics.
We follow the training/testing/validation split of~\cite{gkioxari2019meshrcnn}.

\paragraph{Implementation}\vspace{-4mm}
For the depth prediction module, we follow the original MVSNet~\cite{yao2018mvsnet} implementation.
The output depth dimensions reduces by a factor of 4 to 56$\times$56 from the 224$\times$224 input image.
The number of depth hypotheses is chosen as 48 which offers a balance between accuracy and running/training time efficiency.
These depth hypotheses represent values from $0.1$ m to $1.3$ m at an interval of $25$ mm.
These values were chosen based on the range of depths present in the dataset.

The hierarchical features obtained from "Contrastive Depth Features Extractor" are of total 4800 dimensions for each view.
The aggregated multi-view features are compressed to 480 dimensional after applying attentive feature pooling.
5 attention heads are used for merging multi-view features.
The loss function weights are set as $\lambda_{\text{chamfer}}=1$, $\lambda_{\text{normal}}=1.6\times10^{-4}$, $\lambda_{\text{depth}}=0.1$, $\lambda_{\text{contrastive}}=0.001$ and $\lambda_{\text{voxel}}=1$.
Two settings of $\lambda_{\text{edge}}$ were used, $\lambda_{\text{edge}}=0$ (referred as \emph{Best}) which gives better quantitative results and $\lambda_{\text{edge}}=0.2$ (referred as \emph{Pretty}) which gives better qualitative results.



\paragraph{Training and Runtime}\vspace{-4mm}
The network is optimized using Adam optimizer with a learning rate of $10^{-4}$.
The training is done on 5 Nvidia RTX-2080 GPUs with effective batch size 5.
The depth prediction network (MVSNet) is trained independently for 30 epochs.
Then the whole system is trained for another 40 epochs with the weights of the MVSNet frozen.
Our system is implemented in PyTorch deep learning framework and it takes around 60 hours for training.


\paragraph{Evaluation Metric}\vspace{-4mm}
Following~\cite{wang2018pixel2mesh, wen2019pixel2mesh++}, we use F1-score as our evaluation metric.
The F1-score is the harmonic mean of precision and recall where the precision/recall are calculated by finding the percentage of points in the predicted/ground truth that can find a nearest neighbor from the other within a threshold.
We provide evaluations with two threshold values: $\tau$ and $2\tau$ where $\tau=10^{-4}$  m$^2$.

\vspace{-1mm}
\subsection{Comparison with previous Multi-view Shape Generation Methods}
We quantitatively compare our method against previous works for multi-view shape generation in~\tableref{baseline_comparison} and show the effectiveness of our methods in improving the shape quality. Our method outperforms the state-of-the-art method  Pixel2Mesh++~\cite{wen2019pixel2mesh++} with
a decrease in chamfer distance to ground truth by 34\% and 15\% increase in F1-score at threshold $\tau$.
Note that in~\tableref{baseline_comparison} the same model is trained for all the categories but accuracy on individual categories as well as average over the categories are evaluated.
We provide the chamfer distances in the supplementary material.
\begin{table*}[ht]
\begin{center}
\resizebox{\linewidth}{!}{
\begin{tabular}{|c|cccccc|cccccc|}
    \hline
    \multirow{3}{*}{Category}&
    \multicolumn{6}{c|}{F-score ($\tau$) $\uparrow$}&
    \multicolumn{6}{c|}{F-score ($2\tau$) $\uparrow$}\\
    & \multirow{2}{*}{3D-R2N2}  & \multirow{2}{*}{LSM}  & \multirow{2}{*}{MVP2M}    & \multirow{2}{*}{P2M++} & Ours & \bf{Ours}
    & \multirow{2}{*}{3D-R2N2}  & \multirow{2}{*}{LSM}  & \multirow{2}{*}{MVP2M}    & \multirow{2}{*}{P2M++} & Ours & \bf{Ours} \\
    &                           &                       &                           &                        & (pretty) & \bf{(best)} &
                                &                       &                           &                        & (pretty) & \bf{(best)} \\
    \hline
    Couch       & 45.47 & 43.02 & 53.17 & 57.56 & 71.63 & \bf{73.63}    & 59.97 & 55.49 & 73.24 & 75.33 & 85.28 & \bf{88.24} \\
    Cabinet     & 54.08 & 50.80 & 56.85 & 65.72 & 75.91 & \bf{76.39}    & 64.42 & 60.72 & 76.58 & 81.57 & 87.61 & \bf{88.84} \\
    Bench       & 44.56 & 49.33 & 60.37 & 66.24 & 81.11 & \bf{83.76}    & 62.47 & 65.92 & 75.69 & 79.67 & 90.56 & \bf{92.57} \\
    Chair       & 37.62 & 48.55 & 54.19 & 62.05 & 77.63 & \bf{78.69}    & 54.26 & 64.95 & 72.36 & 77.68 & 88.24 & \bf{90.02} \\
    Monitor     & 36.33 & 43.65 & 53.41 & 60.00 & 74.14 & \bf{76.64}    & 48.65 & 56.33 & 70.63 & 75.42 & 86.04 & \bf{88.89} \\
    Firearm     & 55.72 & 56.14 & 79.67 & 80.74 & 92.92 & \bf{94.32}    & 76.79 & 73.89 & 89.08 & 89.29 & 96.81 & \bf{97.67} \\
    Speaker     & 41.48 & 45.21 & 48.90 & 54.88 & 66.02 & \bf{67.83}    & 52.29 & 56.65 & 68.29 & 71.46 & 79.76 & \bf{82.34} \\
    Lamp        & 32.25 & 45.58 & 50.82 & 62.56 & 72.47 & \bf{75.93}    & 49.38 & 64.76 & 65.72 & 74.00 & 82.00 & \bf{85.33} \\
    Cellphone   & 58.09 & 60.11 & 66.07 & 74.36 & 85.57 & \bf{86.45}    & 69.66 & 71.39 & 82.31 & 86.16 & 93.40 & \bf{94.28} \\
    Plane       & 47.81 & 55.60 & 75.16 & 76.79 & 89.23 & \bf{92.13}    & 70.49 & 76.39 & 86.38 & 86.62 & 94.65 & \bf{96.57} \\
    Table       & 48.78 & 48.61 & 65.95 & 71.89 & 82.37 & \bf{83.68}    & 62.67 & 62.22 & 79.96 & 84.19 & 90.24 & \bf{91.97} \\
    Car         & 59.86 & 51.91 & 67.27 & 68.45 & 77.01 & \bf{80.43}    & 78.31 & 68.20 & 84.64 & 85.19 & 88.99 & \bf{92.33} \\
    Watercraft  & 40.72 & 47.96 & 61.85 & 62.99 & 75.52 & \bf{80.48}    & 63.59 & 66.95 & 77.49 & 77.32 & 86.77 & \bf{90.35} \\
    \hline
    Mean        & 46.37 & 49.73 & 61.05 & 66.48 & 78.58 & \bf{80.80}    & 62.53 & 64.91 & 77.10 & 80.30 & 88.49 & \bf{90.72} \\
    \hline
\end{tabular}}
\end{center}
\vspace{-4mm}
\caption{
    \textbf{Qualitative comparison} against state-of-the-art multi-view shape generation methods. We report F-score on each semantic category along with the mean over all categories using two thresholds $\tau$ and $2\tau$ for nearest neighbor match where ${\tau}$=$10^{-4}$ m$^2$.
}
\label{table:baseline_comparison}
\end{table*}

We also provide visual results for qualitative assessment of the generated shapes by our \emph{Pretty} model in~\figref{qualitative_evaluation} which shows that it is able to more accurately predict topologically diverse shapes.


\subsection{Ablation studies}


\begin{table}[ht]
\begin{center}
\footnotesize
\begin{tabular}{ l c c }
\toprule[1pt]
 &F1-$\tau$ &F1-2$\tau$   \\ \hline
(1) Naive multi-view Mesh R-CNN \qquad \qquad  \qquad  \qquad  \qquad & 72.74 & 84.99 \\
(2) + Multi-view voxel grid prediction & 76.97 & 88.24 \\
(3) + Contrastive depth input             & 79.63 & 90.10 \\
(4) + Attention pooling   & 79.82    & 90.18  \\
(5) \bf{+ Contrastive depth loss (final model)}  & \textbf{80.80} & \textbf{90.72}\\
(6) Using GT depth (final model)                 & \textbf{84.58} & \textbf{92.86} \\
\bottomrule[1pt]
\end{tabular}
\end{center}
\vspace{-4mm}
\caption{
    \textbf{Comparison of shape generation accuracy with different settings} of additional contrastive depth losses, multi-view voxel grid prediction and feature pooling.}
\label{table:ablation_study}
\end{table}


\paragraph{Accuracy with different settings}
\tableref{ablation_study} shows the contribution of different components towards the final accuracy. Naively extending the single-view Mesh R-CNN~\cite{gkioxari2019meshrcnn} to multiple views using statistical feature pooling~\cite{wen2019pixel2mesh++} for mesh refinement (row 1) gives an F1-score of 72.74\% for threshold $\tau$ which is 6.26\% improvement over Pixel2Mesh++.
We further extend the above method with our probabilistic multi-view voxel grid prediction in row 2 and get a 4.23\% improvement.

In row 3 of~\tableref{ablation_study} we use our contrastive depth features instead of RGB features for mesh refinement and get 2.7\% improvement.
We then replace the statistical feature pooling with the proposed attention method and get 0.19\% improvement.
The improvement is not significant on our final architecture but we found the attention to perform better on more light-weight architectures and since it leads to lower memory consumption than statistical feature pooling, we decide to keep it.
We also evaluate the effect of using additional regularization from contrastive depth losses: rendered depth vs predicted depth in the 5th rows of which improves the score by 0.98\%.
In row 6 we use ground truth instead of predicted depths on our final model which gives the upper bound on our mesh prediction accuracy in relation to the depth prediction accuracy as 84.58\%.

\begin{table}[ht]
\begin{center}
\footnotesize
\begin{tabular}{ l c c }
\toprule[1pt]
 &F1-$\tau$ &F1-2$\tau$   \\ \hline
(1) P2M++ & 66.48 & 80.30 \\
(2) Sphere initialization & 73.78 & 85.49 \\
(3) Back-projected predicted depth initialization & 75.62 & 86.53 \\
(4) \bf{Proposed model}  & \textbf{80.80} & \textbf{90.72}\\
\bottomrule[1pt]
\end{tabular}
\end{center}
\vspace{-4mm}
\caption{
    \textbf{Comparison of shape generation accuracy with different initial shapes}: sphere, back-projected depth and the proposed model (CNN-based multi-view voxel grid prediction)}
\label{table:initial_shape}
\end{table}

\paragraph{Initial Shape}\vspace{-4mm}
\tableref{initial_shape} shows the F1-scores when using different methods for generating initial shapes for GCN refinement.
All the methods (except P2M++) use contrastive depth features and loss along with attention-based feature pooling.
When a unit level-4 ico-sphere is used as the initial shape (row 2), the score is F1-$\tau$ 73.78\%.
Using back-projected predicted depth maps as initial shape (row 3), the scores increases to 75.62\%.
Here, the back-projected points are transformed to 48 $\times$ 48 $\times$ 48 voxel grid
which can be done differentiably and efficiently using GPU based k-nearest neighbor calculation to determine the occupancy of each voxel.
The proposed model which uses CNN-based voxel grid prediction (row 4) has the highest score of 80.8\%.
The difference in score when using back-projected depth vs CNN predicted shape is due to the incomplete shapes the predicted depth give due to occlusion along with depth error due to arbitrary baseline between input images which is challenging for MVSNet-based depth predictor.

\paragraph{Contrastive Depth Feature Extraction}\vspace{-4mm}
We evaluate several methods for contrastive feature extraction (\subsecref{contrastive_depth_feature_extraction}). These methods are
1) \emph{Input Concatenation}: using the concatenated rendered and predicted depth maps as input to the CNN feature extractor,
2) \emph{Input Difference}: using the difference of the two depth maps as input to CNN,
3) \emph{Feature Concatenation}: concatenating features from rendered and predicted depths extracted by shared CNN,
4) \emph{Feature Difference}: using difference of the features from the two depth maps extracted by shared CNN, and
5) \emph{Predicted depth only}: using the CNN features from the predicted depths only.
6) \emph{Rendered depth only}: using the CNN features from the rendered depths only.
The quantitative results are summarized in Table~\ref{table:contrastive_feature_extraction} and shows that \emph{Input Concatenation} method produces better results than other formulations.

\begin{table}[ht]
\begin{center}
\footnotesize
\begin{tabular}{ l c c }
\toprule[1pt]
 &F1-$\tau$ &F1-2$\tau$ \\ \hline
(1) Input Concatenation \qquad \qquad  \qquad  \qquad  \qquad  & \textbf{80.80} & \textbf{90.72} \\
(2) Input Difference & 80.41 & 90.54 \\
(3) Feature Concatenation   & 80.45 & 90.54 \\
(4) Feature Difference & 80.30 & 90.40 \\
(5) Predicted Depth only & 79.40 & 89.95 \\
(6) Rendered Depth only & 78.20 & 88.90 \\
\bottomrule[1pt]
\end{tabular}
\end{center}
\vspace{-4mm}
\caption{\textbf{Comparisons of different contrastive depth formulations}. In 1st and 2nd rows, concatenation and difference of the rendered and predicted depths are fed to CNN feature extractor while in 3rd and 4th rows, concatenation and difference of the CNN features from the depths is used for mesh refinement. 5 uses features from predicted depths only while 6 uses features from rendered depths only.}
\label{table:contrastive_feature_extraction}
\end{table}





\paragraph{Number of Views}\vspace{-4mm}
We test the performance of our framework with respect to the number of views.
\tableref{number_of_input_views} shows that the accuracy of our method increases as we increase the number of input views for training.

\tableref{number_of_test_views} shows the results when using different number of views during testing on our model trained with 3 views
which indicates that increasing the number of views during testing does not improve the accuracy while decreasing the number of views can cause a significant drop in accuracy.
\begin{table}[ht]
\noindent \scriptsize \footnotesize
\begin{minipage}[t]{0.5\textwidth}
\centering
\begin{tabular}{ l | c c c c c }
    \hline
    Metric & 2 & 3 & 4 & 5 & 6\\
    \hline
    F1-$\tau$  & 73.60 & 80.80 & 82.61 & 83.76 & 84.25 \\
    F1-$2\tau$ & 85.80 & 90.72 & 91.78 & 92.73 & 93.14 \\
    \hline
\end{tabular}
\caption{
    \textbf{Accuracy w.r.t the number of views during training}.
    The evaluation was performed on the same number of views as training.
}
\label{table:number_of_input_views}
\end{minipage}
\hspace{0.1cm}
\noindent \scriptsize \footnotesize
\begin{minipage}[t]{0.5\textwidth}
\centering
\begin{tabular}{ l | c c c c c }
    \hline
    Metric & 2 & 3 & 4 & 5 & 6 \\
    \hline
    F1-$\tau$   & 72.46 & 80.80 & 80.98 & 80.94 & 80.85 \\
    F1-$2\tau$  & 84.49 & 90.72 & 91.03 & 91.16 & 91.20 \\
    \hline
\end{tabular}
\caption{
    \textbf{Accuracy w.r.t the number of views during testing}.
    The same model trained with 3 views was used in all of the cases.
}
\label{table:number_of_test_views}
\end{minipage}
\end{table}

\begin{figure}[th!]
    \begin{center}
        \includegraphics[width=\linewidth]{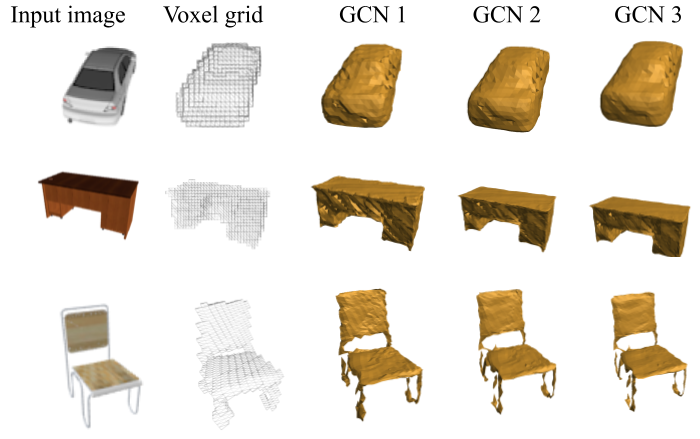}
    \end{center}
        \caption{Predicted shapes at different stages in the pipeline.}
        \label{fig:gcn_stages}
\end{figure}

\paragraph{Accuracy at Different GCN Stages}\vspace{-2mm}
We analyze the accuracy of meshes at different GCN stages in~\tableref{gcn_stages} and~\figref{gcn_stages}. The results validate that our method produces the meshes in a coarse-to-fine manner and multiple GCN refinements improve the mesh quality.
\begin{table}[ht!]
\begin{center}
\begin{tabular}{c | c c c c}
    \hline
    Metric & Cubified & 1 & 2 & 3 \\
    \hline
    F1-$\tau$   & 31.48 & 76.78 & 79.88 & 80.80 \\
    F1-$2\tau$  & 44.40 & 88.32 & 90.19 & 90.72 \\
    \hline
\end{tabular}
\end{center}
\caption{
    Accuracy at different GCN stages. 1, 2 and 3 indicate the performance at the corresponding graph convolution blocks while \emph{Cubified} is for the cubified voxel grid used as input for the first GCN block.
}
\label{table:gcn_stages}
\end{table}

\vspace{-2mm}
\paragraph{Generalization Capability}
We conduct experiments to evaluate the generalization capability of our system across the semantic categories. We train our model with only 12 out of the 13 categories and test on the category that was left out. \tableref{train_except_one_category} shows that the accuracy generally does not decrease significantly when compared with the model that was trained on all 13 categories, especially when using $2\tau$ threshold for F-score.
\begin{table}[ht]
\begin{center}
\footnotesize
\begin{tabular}{| c | c c | c c|}
    \hline
    \multirow{2}{*}{Category}&
    \multicolumn{2}{c|}{F-score ($\tau$) $\uparrow$}&
    \multicolumn{2}{c|}{F-score ($2\tau$) $\uparrow$}\\
    & Excluding & Including
    & Excluding & Including \\
    \hline
    Couch       & 63.29 & 73.63 & 80.79 & 88.24 \\
    Cabinet     & 68.26 & 76.39 & 83.10 & 88.84 \\
    Bench       & 76.08 & 83.76 & 87.42 & 92.57 \\
    Chair       & 60.60 & 78.69 & 75.93 & 90.02 \\
    Monitor     & 67.26 & 76.64 & 81.57 & 88.89 \\
    Firearm     & 78.59 & 94.32 & 86.28 & 97.67 \\
    Speaker     & 62.39 & 67.83 & 77.77 & 82.34 \\
    Lamp        & 63.50 & 75.93 & 74.66 & 85.33 \\
    Cellphone   & 67.24 & 86.45 & 80.54 & 94.28 \\
    Plane       & 57.48 & 92.13 & 67.27 & 96.57 \\
    Table       & 76.41 & 83.68 & 86.86 & 91.97 \\
    Car         & 59.08 & 80.43 & 75.58 & 92.33 \\
    Watercraft  & 64.97 & 80.48 & 78.95 & 90.35 \\
    \hline
\end{tabular}
\end{center}
\caption{
    \textbf{Accuracy when a category is excluded} during training and evaluation is performed on the category to verify how well training on other categories generalizes to the excluded category.
}
\label{table:train_except_one_category}
\end{table}

\section{Conclusion}

We propose a neural network based solution to predict 3D triangle mesh models of objects from images taken from multiple views.
First, we propose a multi-view voxel grid prediction module which probabilistically merges voxel grids predicted from individual input views.
We then cubify the merged voxel grid to triangle mesh and apply graph convolutional networks for further refining the mesh.
The features for the mesh vertices are extracted from contrastive depth input consisting of rendered depths at each refinement stage along with the predicted depths.
The proposed mesh reconstruction method outperforms existing methods with a large margin and is capable of reconstructing objects with more complex topologies.

{\small
\bibliographystyle{ieee_fullname}
\bibliography{egbib}

\begin{thebibliography}{10}\itemsep=-1pt

\bibitem{agarwal2011building}
Sameer Agarwal, Yasutaka Furukawa, Noah Snavely, Ian Simon, Brian Curless,
  Steven~M Seitz, and Richard Szeliski.
\newblock Building rome in a day.
\newblock {\em Communications of the ACM}, 54(10):105--112, 2011.

\bibitem{cadena2016pastslam}
Cesar Cadena, Luca Carlone, Henry Carrillo, Yasir Latif, Davide Scaramuzza,
  Jos{\'e} Neira, Ian Reid, and John~J Leonard.
\newblock Past, present, and future of simultaneous localization and mapping:
  Toward the robust-perception age.
\newblock {\em IEEE Transactions on robotics}, 32(6):1309--1332, 2016.

\bibitem{chang2015shapenet}
Angel~X Chang, Thomas Funkhouser, Leonidas Guibas, Pat Hanrahan, Qixing Huang,
  Zimo Li, Silvio Savarese, Manolis Savva, Shuran Song, Hao Su, et~al.
\newblock Shapenet: An information-rich 3d model repository.
\newblock {\em arXiv preprint arXiv:1512.03012}, 2015.

\bibitem{chen2019point}
Rui Chen, Songfang Han, Jing Xu, and Hao Su.
\newblock Point-based multi-view stereo network.
\newblock In {\em Proceedings of the IEEE International Conference on Computer
  Vision}, pages 1538--1547, 2019.

\bibitem{3dr2n2}
Christopher~B Choy, Danfei Xu, JunYoung Gwak, Kevin Chen, and Silvio Savarese.
\newblock 3d-r2n2: A unified approach for single and multi-view 3d object
  reconstruction.
\newblock In {\em European conference on computer vision}, pages 628--644.
  Springer, 2016.

\bibitem{cui2017hsfm}
Hainan Cui, Xiang Gao, Shuhan Shen, and Zhanyi Hu.
\newblock Hsfm: Hybrid structure-from-motion.
\newblock In {\em Proceedings of the IEEE Conference on Computer Vision and
  Pattern Recognition}, pages 1212--1221, 2017.

\bibitem{cui2015global}
Zhaopeng Cui and Ping Tan.
\newblock Global structure-from-motion by similarity averaging.
\newblock In {\em Proceedings of the IEEE International Conference on Computer
  Vision}, pages 864--872, 2015.

\bibitem{durou2008numerical}
Jean-Denis Durou, Maurizio Falcone, and Manuela Sagona.
\newblock Numerical methods for shape-from-shading: A new survey with
  benchmarks.
\newblock {\em Computer Vision and Image Understanding}, 109(1):22--43, 2008.

\bibitem{engel2014lsd}
Jakob Engel, Thomas Sch{\"o}ps, and Daniel Cremers.
\newblock Lsd-slam: Large-scale direct monocular slam.
\newblock In {\em European conference on computer vision}, pages 834--849.
  Springer, 2014.

\bibitem{fan2017point}
Haoqiang Fan, Hao Su, and Leonidas~J Guibas.
\newblock A point set generation network for 3d object reconstruction from a
  single image.
\newblock In {\em Proceedings of the IEEE conference on computer vision and
  pattern recognition}, pages 605--613, 2017.

\bibitem{favaro2005geometric}
Paolo Favaro and Stefano Soatto.
\newblock A geometric approach to shape from defocus.
\newblock {\em IEEE Transactions on Pattern Analysis and Machine Intelligence},
  27(3):406--417, 2005.

\bibitem{furukawa2009accurate}
Yasutaka Furukawa and Jean Ponce.
\newblock Accurate, dense, and robust multiview stereopsis.
\newblock {\em IEEE transactions on pattern analysis and machine intelligence},
  32(8):1362--1376, 2009.

\bibitem{gkioxari2019meshrcnn}
Georgia Gkioxari, Jitendra Malik, and Justin Johnson.
\newblock Mesh r-cnn.
\newblock In {\em Proceedings of the IEEE International Conference on Computer
  Vision}, pages 9785--9795, 2019.

\bibitem{grisetti2007improved}
Giorgio Grisetti, Cyrill Stachniss, and Wolfram Burgard.
\newblock Improved techniques for grid mapping with rao-blackwellized particle
  filters.
\newblock {\em IEEE transactions on Robotics}, 23(1):34--46, 2007.

\bibitem{groueix2018papier}
Thibault Groueix, Matthew Fisher, Vladimir~G Kim, Bryan~C Russell, and Mathieu
  Aubry.
\newblock A papier-m{\^a}ch{\'e} approach to learning 3d surface generation.
\newblock In {\em Proceedings of the IEEE conference on computer vision and
  pattern recognition}, pages 216--224, 2018.

\bibitem{gu2019cascade}
Xiaodong Gu, Zhiwen Fan, Siyu Zhu, Zuozhuo Dai, Feitong Tan, and Ping Tan.
\newblock Cascade cost volume for high-resolution multi-view stereo and stereo
  matching.
\newblock {\em arXiv preprint arXiv:1912.06378}, 2019.

\bibitem{mcrecon2017}
JunYoung Gwak, Christopher~B Choy, Manmohan Chandraker, Animesh Garg, and
  Silvio Savarese.
\newblock Weakly supervised 3d reconstruction with adversarial constraint.
\newblock In {\em 2017 International Conference on 3D Vision (3DV)}, pages
  263--272. IEEE, 2017.

\bibitem{han2020seqxy2seqz}
Zhizhong Han, Guanhui Qiao, Yu-Shen Liu, and Matthias Zwicker.
\newblock Seqxy2seqz: Structure learning for 3d shapes by sequentially
  predicting 1d occupancy segments from 2d coordinates.
\newblock {\em arXiv preprint arXiv:2003.05559}, 2020.

\bibitem{hane2017hierarchical}
Christian H{\"a}ne, Shubham Tulsiani, and Jitendra Malik.
\newblock Hierarchical surface prediction for 3d object reconstruction.
\newblock In {\em 2017 International Conference on 3D Vision (3DV)}, pages
  412--420. IEEE, 2017.

\bibitem{hartmann2017learned_16}
Wilfried Hartmann, Silvano Galliani, Michal Havlena, Luc Van~Gool, and Konrad
  Schindler.
\newblock Learned multi-patch similarity.
\newblock In {\em Proceedings of the IEEE International Conference on Computer
  Vision}, pages 1586--1594, 2017.

\bibitem{deepmvs2018}
Po-Han Huang, Kevin Matzen, Johannes Kopf, Narendra Ahuja, and Jia-Bin Huang.
\newblock Deepmvs: Learning multi-view stereopsis.
\newblock In {\em Proceedings of the IEEE Conference on Computer Vision and
  Pattern Recognition}, pages 2821--2830, 2018.

\bibitem{huang2015single}
Qixing Huang, Hai Wang, and Vladlen Koltun.
\newblock Single-view reconstruction via joint analysis of image and shape
  collections.
\newblock {\em ACM Transactions on Graphics (TOG)}, 34(4):1--10, 2015.

\bibitem{jia2020dv}
Xin Jia, Shourui Yang, Yuxin Peng, Junchao Zhang, and Shengyong Chen.
\newblock Dv-net: Dual-view network for 3d reconstruction by fusing multiple
  sets of gated control point clouds.
\newblock {\em Pattern Recognition Letters}, 131:376--382, 2020.

\bibitem{jin2019drkfs}
Jiongchao Jin, Akshay~Gadi Patil, Zhang Xiong, and Hao Zhang.
\newblock Dr-kfs: A differentiable visual similarity metric for 3d shape
  reconstruction, 2019.

\bibitem{johnston2017scaling}
Adrian Johnston, Ravi Garg, Gustavo Carneiro, Ian Reid, and Anton van~den
  Hengel.
\newblock Scaling cnns for high resolution volumetric reconstruction from a
  single image.
\newblock In {\em Proceedings of the IEEE International Conference on Computer
  Vision Workshops}, pages 939--948, 2017.

\bibitem{kar2017lsm}
Abhishek Kar, Christian H{\"a}ne, and Jitendra Malik.
\newblock Learning a multi-view stereo machine.
\newblock In {\em Advances in neural information processing systems}, pages
  365--376, 2017.

\bibitem{kar2015category}
Abhishek Kar, Shubham Tulsiani, Joao Carreira, and Jitendra Malik.
\newblock Category-specific object reconstruction from a single image.
\newblock In {\em Proceedings of the IEEE conference on computer vision and
  pattern recognition}, pages 1966--1974, 2015.

\bibitem{kato2018renderer}
Hiroharu Kato, Yoshitaka Ushiku, and Tatsuya Harada.
\newblock Neural 3d mesh renderer.
\newblock In {\em The IEEE Conference on Computer Vision and Pattern
  Recognition (CVPR)}, 2018.

\bibitem{konolige1997improved}
Kurt Konolige.
\newblock Improved occupancy grids for map building.
\newblock {\em Autonomous Robots}, 4(4):351--367, 1997.

\bibitem{kurenkov2018deformnet}
Andrey Kurenkov, Jingwei Ji, Animesh Garg, Viraj Mehta, JunYoung Gwak,
  Christopher Choy, and Silvio Savarese.
\newblock Deformnet: Free-form deformation network for 3d shape reconstruction
  from a single image.
\newblock In {\em 2018 IEEE Winter Conference on Applications of Computer
  Vision (WACV)}, pages 858--866. IEEE, 2018.

\bibitem{kutulakos2000theory}
Kiriakos~N Kutulakos and Steven~M Seitz.
\newblock A theory of shape by space carving.
\newblock {\em International journal of computer vision}, 38(3):199--218, 2000.

\bibitem{laina2016deeper}
Iro Laina, Christian Rupprecht, Vasileios Belagiannis, Federico Tombari, and
  Nassir Navab.
\newblock Deeper depth prediction with fully convolutional residual networks.
\newblock In {\em 2016 Fourth international conference on 3D vision (3DV)},
  pages 239--248. IEEE, 2016.

\bibitem{lhuillier2005quasi}
Maxime Lhuillier and Long Quan.
\newblock A quasi-dense approach to surface reconstruction from uncalibrated
  images.
\newblock {\em IEEE transactions on pattern analysis and machine intelligence},
  27(3):418--433, 2005.

\bibitem{lin2019photometric}
Chen-Hsuan Lin, Oliver Wang, Bryan~C Russell, Eli Shechtman, Vladimir~G Kim,
  Matthew Fisher, and Simon Lucey.
\newblock Photometric mesh optimization for video-aligned 3d object
  reconstruction.
\newblock In {\em Proceedings of the IEEE Conference on Computer Vision and
  Pattern Recognition}, pages 969--978, 2019.

\bibitem{liu2019learning}
Shichen Liu, Shunsuke Saito, Weikai Chen, and Hao Li.
\newblock Learning to infer implicit surfaces without 3d supervision.
\newblock In {\em Advances in Neural Information Processing Systems}, pages
  8293--8304, 2019.

\bibitem{liu2019dist}
Shaohui Liu, Yinda Zhang, Songyou Peng, Boxin Shi, Marc Pollefeys, and Zhaopeng
  Cui.
\newblock Dist: Rendering deep implicit signed distance function with
  differentiable sphere tracing.
\newblock {\em arXiv preprint arXiv:1911.13225}, 2019.

\bibitem{luo2019pmvsnet}
Keyang Luo, Tao Guan, Lili Ju, Haipeng Huang, and Yawei Luo.
\newblock P-mvsnet: Learning patch-wise matching confidence aggregation for
  multi-view stereo.
\newblock In {\em Proceedings of the IEEE International Conference on Computer
  Vision}, pages 10452--10461, 2019.

\bibitem{mescheder2019occupancy}
Lars Mescheder, Michael Oechsle, Michael Niemeyer, Sebastian Nowozin, and
  Andreas Geiger.
\newblock Occupancy networks: Learning 3d reconstruction in function space.
\newblock In {\em Proceedings of the IEEE Conference on Computer Vision and
  Pattern Recognition}, pages 4460--4470, 2019.

\bibitem{mur2015orb}
Raul Mur-Artal, Jose Maria~Martinez Montiel, and Juan~D Tardos.
\newblock Orb-slam: a versatile and accurate monocular slam system.
\newblock {\em IEEE transactions on robotics}, 31(5):1147--1163, 2015.

\bibitem{murez2020atlas}
Zak Murez, Tarrence van As, James Bartolozzi, Ayan Sinha, Vijay Badrinarayanan,
  and Andrew Rabinovich.
\newblock Atlas: End-to-end 3d scene reconstruction from posed images.
\newblock {\em arXiv preprint arXiv:2003.10432}, 2020.

\bibitem{pan2019deep}
Junyi Pan, Xiaoguang Han, Weikai Chen, Jiapeng Tang, and Kui Jia.
\newblock Deep mesh reconstruction from single rgb images via topology
  modification networks.
\newblock In {\em Proceedings of the IEEE International Conference on Computer
  Vision}, pages 9964--9973, 2019.

\bibitem{park2019deepsdf}
Jeong~Joon Park, Peter Florence, Julian Straub, Richard Newcombe, and Steven
  Lovegrove.
\newblock Deepsdf: Learning continuous signed distance functions for shape
  representation.
\newblock In {\em Proceedings of the IEEE Conference on Computer Vision and
  Pattern Recognition}, pages 165--174, 2019.

\bibitem{scarselli2008graph}
Franco Scarselli, Marco Gori, Ah~Chung Tsoi, Markus Hagenbuchner, and Gabriele
  Monfardini.
\newblock The graph neural network model.
\newblock {\em IEEE Transactions on Neural Networks}, 20(1):61--80, 2008.

\bibitem{schonberger2016structure}
Johannes~L Schonberger and Jan-Michael Frahm.
\newblock Structure-from-motion revisited.
\newblock In {\em Proceedings of the IEEE Conference on Computer Vision and
  Pattern Recognition}, pages 4104--4113, 2016.

\bibitem{seitz1999photorealistic}
Steven~M Seitz and Charles~R Dyer.
\newblock Photorealistic scene reconstruction by voxel coloring.
\newblock {\em International Journal of Computer Vision}, 35(2):151--173, 1999.

\bibitem{simonyan2014vgg}
Karen Simonyan and Andrew Zisserman.
\newblock Very deep convolutional networks for large-scale image recognition.
\newblock {\em arXiv preprint arXiv:1409.1556}, 2014.

\bibitem{su2014estimating}
Hao Su, Qixing Huang, Niloy~J Mitra, Yangyan Li, and Leonidas Guibas.
\newblock Estimating image depth using shape collections.
\newblock {\em ACM Transactions on Graphics (TOG)}, 33(4):1--11, 2014.

\bibitem{tang2019skeleton}
Jiapeng Tang, Xiaoguang Han, Junyi Pan, Kui Jia, and Xin Tong.
\newblock A skeleton-bridged deep learning approach for generating meshes of
  complex topologies from single rgb images.
\newblock In {\em Proceedings of the IEEE Conference on Computer Vision and
  Pattern Recognition}, pages 4541--4550, 2019.

\bibitem{tulsiani2017multi}
Shubham Tulsiani, Tinghui Zhou, Alexei~A Efros, and Jitendra Malik.
\newblock Multi-view supervision for single-view reconstruction via
  differentiable ray consistency.
\newblock In {\em Proceedings of the IEEE conference on computer vision and
  pattern recognition}, pages 2626--2634, 2017.

\bibitem{vaswani2017attention}
Ashish Vaswani, Noam Shazeer, Niki Parmar, Jakob Uszkoreit, Llion Jones,
  Aidan~N Gomez, {\L}ukasz Kaiser, and Illia Polosukhin.
\newblock Attention is all you need.
\newblock In {\em Advances in neural information processing systems}, pages
  5998--6008, 2017.

\bibitem{wang2018pixel2mesh}
Nanyang Wang, Yinda Zhang, Zhuwen Li, Yanwei Fu, Wei Liu, and Yu-Gang Jiang.
\newblock Pixel2mesh: Generating 3d mesh models from single rgb images.
\newblock In {\em Proceedings of the European Conference on Computer Vision
  (ECCV)}, pages 52--67, 2018.

\bibitem{wen2019pixel2mesh++}
Chao Wen, Yinda Zhang, Zhuwen Li, and Yanwei Fu.
\newblock Pixel2mesh++: Multi-view 3d mesh generation via deformation.
\newblock In {\em Proceedings of the IEEE International Conference on Computer
  Vision}, pages 1042--1051, 2019.

\bibitem{whelan2015elasticfusion}
Thomas Whelan, Stefan Leutenegger, R Salas-Moreno, Ben Glocker, and Andrew
  Davison.
\newblock Elasticfusion: Dense slam without a pose graph.
\newblock Robotics: Science and Systems, 2015.

\bibitem{yan2016perspective}
Xinchen Yan, Jimei Yang, Ersin Yumer, Yijie Guo, and Honglak Lee.
\newblock Perspective transformer nets: Learning single-view 3d object
  reconstruction without 3d supervision.
\newblock In {\em Advances in neural information processing systems}, pages
  1696--1704, 2016.

\bibitem{yang2018foldingnet}
Yaoqing Yang, Chen Feng, Yiru Shen, and Dong Tian.
\newblock Foldingnet: Point cloud auto-encoder via deep grid deformation.
\newblock In {\em Proceedings of the IEEE Conference on Computer Vision and
  Pattern Recognition}, pages 206--215, 2018.

\bibitem{yao2018mvsnet}
Yao Yao, Zixin Luo, Shiwei Li, Tian Fang, and Long Quan.
\newblock Mvsnet: Depth inference for unstructured multi-view stereo.
\newblock In {\em Proceedings of the European Conference on Computer Vision
  (ECCV)}, pages 767--783, 2018.

\bibitem{yao2019recurrent}
Yao Yao, Zixin Luo, Shiwei Li, Tianwei Shen, Tian Fang, and Long Quan.
\newblock Recurrent mvsnet for high-resolution multi-view stereo depth
  inference.
\newblock In {\em Proceedings of the IEEE Conference on Computer Vision and
  Pattern Recognition}, pages 5525--5534, 2019.

\bibitem{yao2020front2back}
Yuan Yao, Nico Schertler, Enrique Rosales, Helge Rhodin, Leonid Sigal, and Alla
  Sheffer.
\newblock Front2back: Single view 3d shape reconstruction via front to back
  prediction.
\newblock In {\em Proceedings of the IEEE/CVF Conference on Computer Vision and
  Pattern Recognition}, pages 531--540, 2020.

\bibitem{zhang1999shape}
Ruo Zhang, Ping-Sing Tsai, James~Edwin Cryer, and Mubarak Shah.
\newblock Shape-from-shading: a survey.
\newblock {\em IEEE transactions on pattern analysis and machine intelligence},
  21(8):690--706, 1999.

\end{thebibliography}
}

\appendix
\section*{Supplementary material}
\section{Network architecture}
\subsection{MVSNet architecture}
\label{subsec:mvsnet}

\begin{figure*}[ht]
    \begin{center}
        \includegraphics[width=\linewidth]{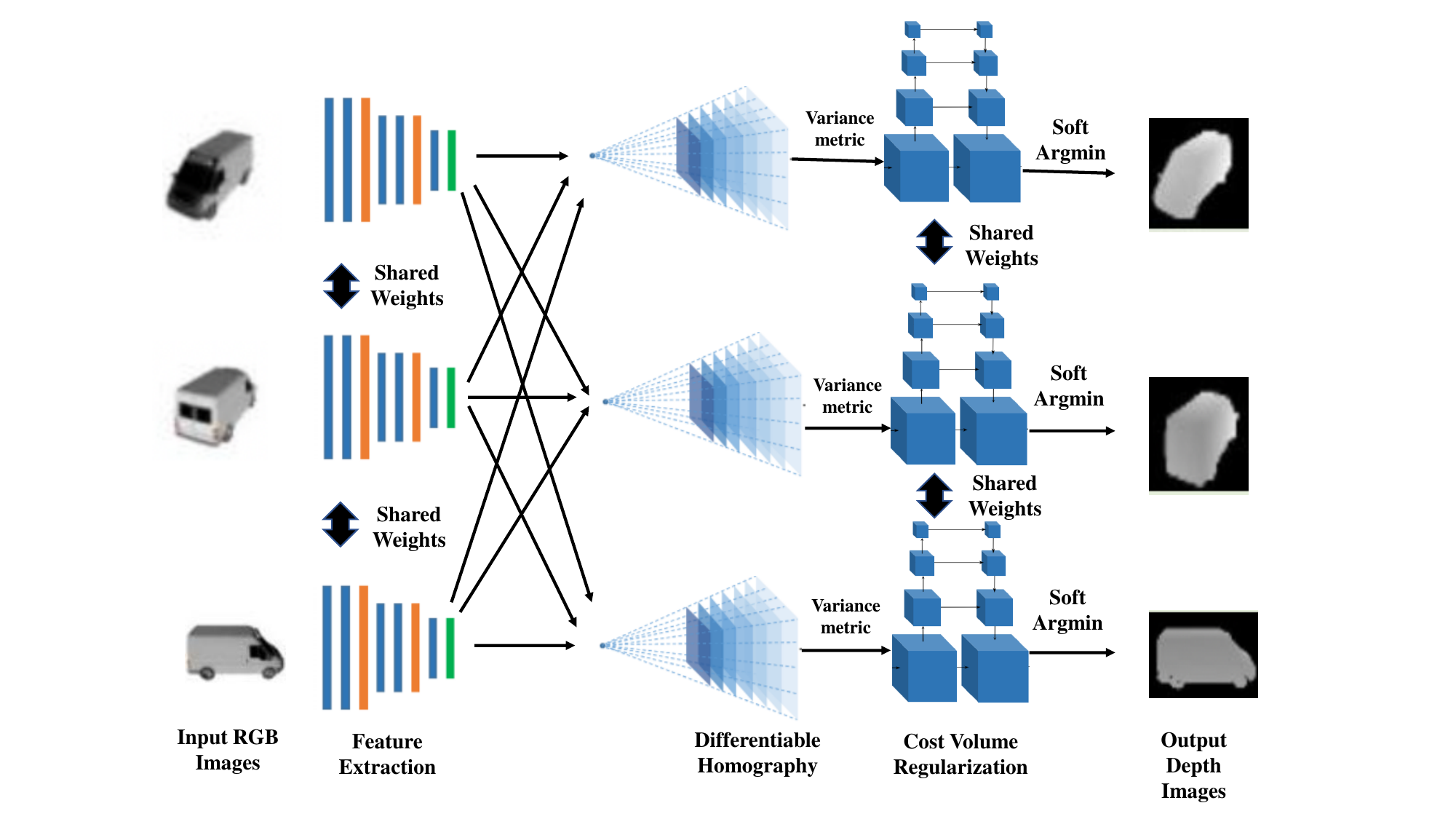}
    \end{center}
    \caption{Depth prediction network (MVSNet) architecture}
    \label{fig:mvsnet_architecture}
\end{figure*}

Our depth prediction module is based on MVSNet~\cite{yao2018mvsnet} which constructs a regularized 3D cost volumes
to estimate the depth map of the reference view.
Here, we extent MVSNet to predict the depth maps of all views instead of only the reference view.
This is achieved by transforming the feature volumes to each view's coordinate frame using homography warping
and applying identical cost volume regularization and depth regression on each view.
This allows the reuse of pre-regularization feature volumes for efficient multi-view depth prediction invariant to the order of input images.
\figref{mvsnet_architecture} shows the architecture of the our depth estimation module.

\subsection{Probabilistic Occupancy Grid Merging}
We use single-view voxel prediction network from~\cite{gkioxari2019meshrcnn} to predict predicts voxel grids for each of the input images in their respective local coordinate frames.
The occupancy grids are transformed to global frame (which is set to the coordinate frame of the first image)
by finding the equivalent global grid values in the local grids after applying bilinear interpolation on the closest matches.
The voxel grids in global coordinates are then probabilistically merged according to Subsection 3.1 of the main submission.

\section{Experiments}

We quantitatively compare our method against previous works for multi-view shape generation in~\tableref{baseline_comparison_cd} and show effectiveness of our proposed shape generation methods in improving shape quality. Our method outperforms the state-of-the-art method  Pixel2Mesh++~\cite{wen2019pixel2mesh++} with
decrease in chamfer distance to ground truth by 34\%, which shows the effectiveness of our proposed method.
Note that in~\tableref{baseline_comparison_cd} same model is trained for all the categories but accuracy on individual categories as well as average over all the categories are evaluated.

\begin{table}[ht]
\begin{center}
\resizebox{\linewidth}{!}{
\begin{tabular}{|c|ccccc|}
    \hline
    \multirow{2}{*}{Category}&
    \multicolumn{5}{c|}{Chamfer Distance (CD) $\downarrow$} \\
    & 3D-R2N2 & LSM & MVP2M & P2M++ & \bf{Ours} \\
    \hline
    Couch       & 0.806 & 0.730 & 0.534 & 0.439 & \bf{0.220} \\
    Cabinet     & 0.613 & 0.634 & 0.488 & 0.337 & \bf{0.230} \\
    Bench       & 1.362 & 0.572 & 0.591 & 0.549 & \bf{0.159} \\
    Chair       & 1.534 & 0.495 & 0.583 & 0.461 & \bf{0.201} \\
    Monitor     & 1.465 & 0.592 & 0.658 & 0.566 & \bf{0.217} \\
    Firearm     & 0.432 & 0.385 & 0.305 & 0.305 & \bf{0.123} \\
    Speaker     & 1.443 & 0.767 & 0.745 & 0.635 & \bf{0.402} \\
    Lamp        & 6.780 & 1.768 & 0.980 & 1.135 & \bf{0.755} \\
    Cellphone   & 1.161 & 0.362 & 0.445 & 0.325 & \bf{0.138} \\
    Plane       & 0.854 & 0.496 & 0.403 & 0.422 & \bf{0.084} \\
    Table       & 1.243 & 0.994 & 0.511 & 0.388 & \bf{0.181} \\
    Car         & 0.358 & 0.326 & 0.321 & 0.249 & \bf{0.165} \\
    Watercraft  & 0.869 & 0.509 & 0.463 & 0.508 & \bf{0.175} \\
    \hline
    Mean        & 1.455 & 0.664 & 0.541 & 0.486 & \bf{0.211} \\
    \hline
\end{tabular}
}
\end{center}
\caption{
    \textbf{Qualitative comparison} against state-of-the-art multi-view shape generation methods. Following~\cite{wen2019pixel2mesh++}, we report Chamfer Distance in $m^2 \times 1000$ from ground truth for different methods. Note that same model is trained for all the categories but accuracy on individual categories as well as average over all the categories are evaluated.
}
\label{table:baseline_comparison_cd}
\end{table}

\subsection{Ablation studies}

\vspace{-2mm}
\paragraph{Coarse Shape Generation}
We conduct comparisons on voxel grid predicted from our proposed probabilistically merged voxel grids against single view method~\cite{gkioxari2019meshrcnn}.
As is shown in~\tableref{multiview_voxel_accuracy}, the accuracy of the initial shape generated from probabilistically merged voxel grid is higher than that from individual views.

\begin{table}[ht]
\noindent \scriptsize \footnotesize
\begin{minipage}[t]{0.5\textwidth}
\centering
\begin{tabular}{c | c c}
    \hline
    Metric      & Single-view & Multi-view \\
    \hline
    F1-$\tau$   & 25.19 & 31.27 \\
    F1-$2\tau$  & 36.75 & 44.46 \\
    \hline
\end{tabular}
\caption{
    \textbf{Accuracy of predicted voxel grids} from single-view prediction compared against the proposed probabilistically merged multi-view voxel grids. The voxel branch was trained separately without the mesh refinement and evaluation was performed on the cubified voxel grids. We use three views for probabilistic grid merging.
}
\label{table:multiview_voxel_accuracy}
\end{minipage}
\hspace{0.1cm}
\noindent \scriptsize \footnotesize
\begin{minipage}[t]{0.5\textwidth}
\centering
\begin{tabular}{c | c c c c}
    \hline
    Metric & Cubified & Stage-1 & Stage-2 & Stage-3 \\
    \hline
    F1-$\tau$   & 31.48 & 76.78 & 79.88 & 80.80  \\
    F1-$2\tau$  & 44.40 & 88.32 & 90.19 & 90.72  \\
    \hline
\end{tabular}
\caption{
    \textbf{Accuracy of the refined meshes at different GCN stages}. 1, 2 and 3 indicate the performance at the corresponding graph convolution blocks while \emph{Cubified} is for the cubified voxel grids used as input for the first GCN block. All the stages, including the voxel prediction, were trained jointly and hence the accuracy of voxel predictions varies from that in~\tableref{multiview_voxel_accuracy}.
}
\label{table:gcn_stages}
\end{minipage}
\end{table}

\vspace{-2mm}
\paragraph{Resolution of Depth Prediction}
We conduct experiments using different numbers of depth hypotheses in our depth prediction network (\subsecref{mvsnet}), producing depth values at different resolutions.
A higher number of depth hypothesis means finer resolution of the predicted depths.
The quantitative results with different hypothesis numbers are summarized in~\tableref{depth_resolution}. We set depth hypothesis as $48$ for our final architecture which is equivalent to the resolution of $25$ mm.
We observe that the mesh accuracy remain relatively unchanged if we predict depths at finer resolutions.
\begin{table}[ht]
\begin{center}
\footnotesize
\begin{tabular}{l | c c c c}
    \hline
    Metric & 24 & 48 & 72 & 96  \\
    \hline
    F1-$\tau$  & 80.29 & 80.80 & 80.69 & 80.34   \\
    F1-$2\tau$ & 90.43 & 90.72 & 90.74 & 90.47   \\
    \hline
\end{tabular}
\end{center}
\caption{
    \textbf{Accuracy w.r.t the number of depth hypothesis}. A higher number of depth hypothesis increases the resolution of predicted depth values at the expense of higher memory requirement. The range of depths for all the models are same and based on the minimum/maximum depth in the ShapeNet~\cite{chang2015shapenet} dataset.
}
\label{table:depth_resolution}
\end{table}

\section{Attention Weights visualization}
\begin{figure*}[t]
\begin{center}
\includegraphics[width=\linewidth]{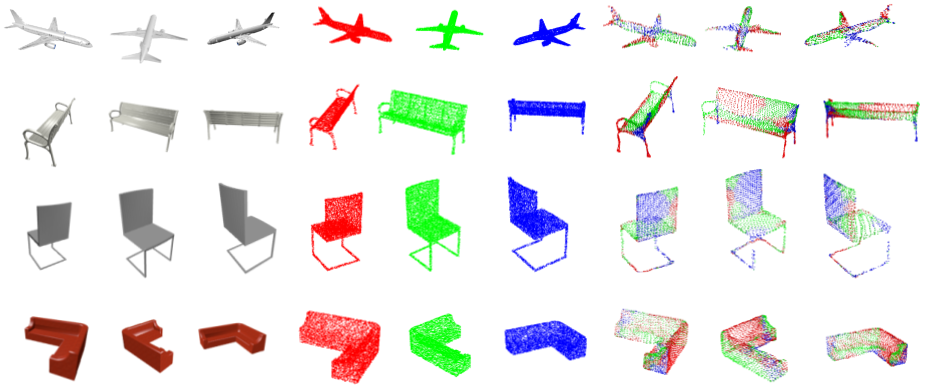}
\end{center}
    \caption{
        \textbf{Attention weights visualization.}
        From left to right: input images from 3 viewpoints, corresponding ground truth point clouds color-coded by their view order and the predicted mesh vertices color-coded by the attention weights of the views.
        Only the view with maximum attention weight is visualized for each predicted points for clarity.
    }
\label{fig:attention_weights}
\end{figure*}

We visualize the learned attention weights (average of each attention heads) in~\figref{attention_weights} where we can observe that the attention weights roughly takes into account the visibility/occlusion information from each view.

\section{Best vs Pretty models}

\begin{figure*}[h]
    \begin{center}
        \includegraphics[width=\linewidth]{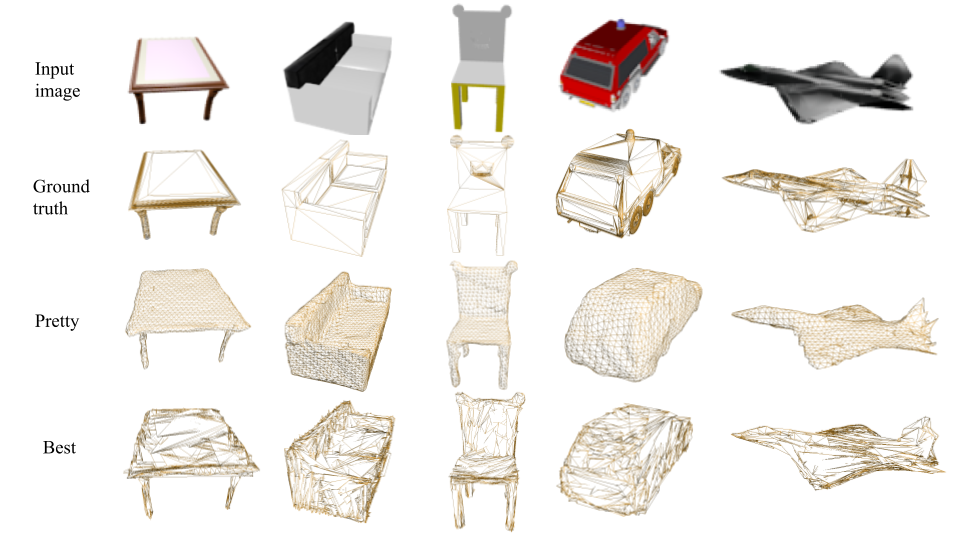}
    \end{center}
    \vspace{-4mm}
        \caption{Qualitative evaluation: best vs pretty wireframe models. The best models while being preferred by the evaluation metrics lead to degenerate meshes, with irregularly sized faces and self-intersections}
        \vspace{-4mm}
        \label{fig:best_vs_pretty}
\end{figure*}

We provide qualitative comparison between the our models trained with \emph{best} and \emph{pretty} configurations in~\figref{best_vs_pretty}.
The \emph{best} configuration refers to our model trained without edge regularization while \emph{pretty} refers to the model trained with the regularization (Subsection 4.1 of the main submission).
We observe that without the regularization we get higher score on our evaluation metrics but get degenerate meshes with self-intersections and irregularly sized faces.

\newpage
\section{Failure Cases}

\begin{figure*}[h!]
    \begin{center}
        \includegraphics[width=\linewidth]{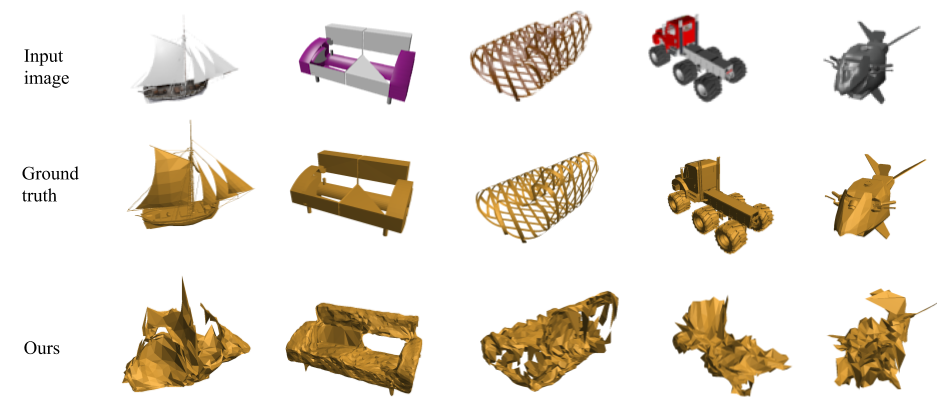}
    \end{center}
    \vspace{-4mm}
        \caption{Failure Cases. Our system can struggle to roughly reconstruct shapes with very complex topology while some fine topology of the mesh is missing.}
        \vspace{-4mm}
        \label{fig:failure_cases}
\end{figure*}

Some failure cases of our model (with pretty setting) are shown in~\figref{failure_cases}.
We notice that the rough topology of the mesh is recovered while we failed to reconstruct the fine topology.
We can regard the recovery from wrong initial topology as a promising future work.

\end{document}